\documentclass[sigconf]{acmart}

\usepackage{balance}

\def\BibTeX{{\rm B\kern-.05em{\sc i\kern-.025em b}\kern-.08emT\kern-.1667em\lower.7ex\hbox{E}\kern-.125emX}}

\copyrightyear{2019}
\acmYear{2019}
\acmConference[MM '19]{Proceedings of the 27th ACM International Conference on
Multimedia}{October 21--25, 2019}{Nice, France}
\acmBooktitle{Proceedings of the 27th ACM International Conference on Multimedia (MM '19),
October 21--25, 2019, Nice, France}
\acmPrice{15.00}
\acmDOI{10.1145/3343031.3350989}
\acmISBN{978-1-4503-6889-6/19/10}

\usepackage{array,multirow}
\usepackage{algorithm}
\usepackage{algorithmic}

\def\ie{{\em i.e.}}
\def\etal{{\em et al.}}

\begin{document}

%
\title{Data Priming Network for Automatic Check-Out}

%
\author{Congcong Li$^{1\ast}$, Dawei Du$^{2\ast}$, Libo Zhang$^{3,4\dagger}$, Tiejian Luo$^{1\dagger}$, Yanjun Wu$^{3,4}$, Qi Tian$^5$, Longyin Wen$^6$, Siwei Lyu$^2$}
\affiliation{%
  \institution{$^1$University of Chinese Academy of Sciences, Beijing, China\\
               $^2$University at Albany, State University of New York, Albany, NY, USA\\
               $^3$Institute of Software Chinese Academy of Sciences, Beijing, China\\
               $^4$State Key Laboratory of Computer Science, ISCAS, Beijing, China\\
               $^5$Huawei Noah's Ark Lab, China\\
               $^6$JD Digits, Mountain View, CA, USA.
               }
}

\thanks{$^\ast$Both authors contributed equally to this research.\\
        $^\dagger$Corresponding author (libo@iscas.ac.cn).}
%
\renewcommand{\shortauthors}{Li and Du, et al.}

%
\begin{abstract}
Automatic Check-Out (ACO) receives increased interests in recent years. An important component of the ACO system is the visual item counting, which recognizes the categories and counts of the items chosen by the customers. However, the training of such a system is challenged by the domain adaptation problem, in which the training data are images from isolated items while the testing images are for collections of items. Existing methods solve this problem with data augmentation using synthesized images, but the image synthesis leads to unreal images that affect the training process. In this paper, we propose a new data priming method to solve the domain adaptation problem. Specifically, we first use pre-augmentation data priming, in which we remove distracting background from the training images using the coarse-to-fine strategy and select images with realistic view angles by the pose pruning method. In the post-augmentation step, we train a data priming network using detection and counting collaborative learning, and select more reliable images from testing data to fine-tune the final visual item tallying network. Experiments on the large scale Retail Product Checkout (RPC) dataset demonstrate the superiority of the proposed method, \ie, we achieve $80.51\%$ checkout accuracy compared with $56.68\%$ of the baseline methods. The source codes can be found in \url{https://isrc.iscas.ac.cn/gitlab/research/acm-mm-2019-ACO}.
\end{abstract}

%
%
%

%

%

%
\maketitle

\section{Introduction}
The recent success of Amazon Go system has invigorated the interests in Automatic Check-Out (ACO) in supermarket and grocery stores. With ACO, customers do not need to put items on the conveyer belt and wait in line for a store assistant to scan them. Instead, they can simply collect the chosen items and an AI-based system will be able to produce the categories and count of these items and automatic process the purchase. Successful ACO system will revolutionize the way we do our shopping and will have significant impact to our daily life in the coming years.
\begin{figure}[t]
\centering
\includegraphics[width=\linewidth]{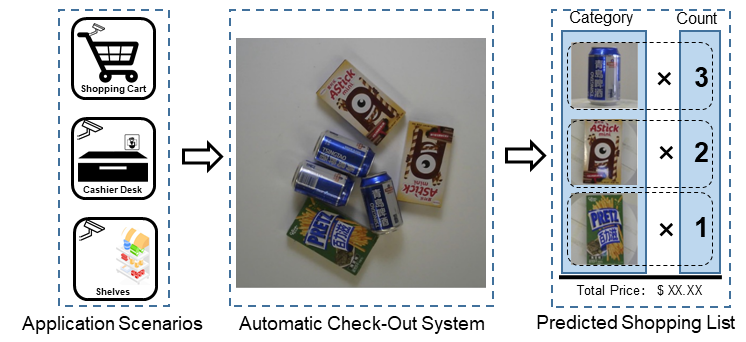}
\caption{The illustration of Automatic Check-Out (ACO) system. It can recognize the categories and counts of the products that the customer puts on the checkout counter, and calculate the corresponding total price. This can be also expanded to other application scenarios such as shopping cart and shelves.}
\label{fig:aco}
\end{figure}

The bedrock of an ACO system is visual item counting that takes images of shopping items as input and generates output as a tally of different categories. With the recent successes of deep learning, deep neural network is a tool of choice for this task. The training of deep neural networks predicates on the availability of large annotated dataset. However, unlike other tasks in computer vision such as object detection and recognition, the training of deep neural network for visual item counting faces a special challenge of domain shift. Specifically, the training data are usually images of individual items under different viewing angles, which is collected using an isolated item sitting on a turntable. As such, the training images may have a distribution different from the images of shopping items piled together over a surface, see Figure~\ref{fig:aco}. The visual item counting algorithm needs to be able to adapt to the difference between the source domain (images of isolated objects) and the target domain (images of collections of objects).

Existing work~\cite{DBLP:journals/corr/abs-1901-07249} attempts to solve this problem with data argumentation. Firstly, images of collections of objects are generated by overlaying individual objects randomly. To improve the realism of the target images, the CycleGAN method~\cite{DBLP:conf/iccv/ZhuPIE17} is used to render realistic shadows and boundaries. However, such a scheme has serious drawbacks. The synthesized testing images have low level of realism due to some unrealistic poses. Besides, there still exists considerable domain shift between training data and testing data.

In this work, we propose a new strategy termed as data priming, to solve the challenging domain adaptation in the visual item counting problem. Instead of simply increasing the data volume by data augmentation as in the previous method~\cite{DBLP:journals/corr/abs-1901-07249}, we improve the relevancy of the augmented data in two steps. In the pre-augmentation data priming step, we extract the foreground region from the training images of isolated objects using the coarse-to-fine saliency detection method. Then, we develop a pose pruning method to choose images only with consistent configurations of the target domain as candidates to generate synthesized images of checked out items with realistic poses. In the post-augmentation data priming step, we construct a data priming network with two heads, one for counting the total number of items and the other for detecting individual objects. Trained on the synthesized images, the data priming network is used to determine the reliability of testing data by detection and counting collaborative learning. Thus reliable testing data is selected to train the visual item tallying network. Experiments on the large-scale Retail Product Checkout (RPC) dataset~\cite{DBLP:journals/corr/abs-1901-07249} demonstrate significant performance improvement of the proposed method compared with the baselines, --- we achieve $80.51\%$ checkout accuracy compared with $56.68\%$ of the baseline method.

The main contributions of this work are three-fold.
\begin{itemize}
\item First, we develop a simple and effective pose pruning method to select synthesized checkout samples with realistic poses for training data.
\item Second, we propose the data priming network to select reliable testing data by detection and counting collaborative learning to guide the training of visual item tallying network.
\item Third, experiments on the RPC dataset shows that our proposed method achieves favorable performance compared to the baselines.
\end{itemize}

\section{Related Work}
In this section, we review previous works that are relevant to the proposed method.

\subsection{Salient Object Detection}
Salient object detection~\cite{DBLP:conf/cvpr/LiY15,DBLP:conf/cvpr/LiuH16,DBLP:journals/tip/HuWZL16,DBLP:conf/mm/TangW17,DBLP:journals/pami/HouCHBTT19} is to segment the main object in the image for pre-processing. Li~\etal~\cite{DBLP:conf/cvpr/LiY15} obtain the saliency map based on the multi-scale features extracted from CNN models. Hu~\etal~\cite{DBLP:journals/tip/HuWZL16} propose a saliency detection method based on the compactness hypothesis that assumes salient regions are more compact than background from the perspectives of both color layout and texture layout. Liu~\etal~\cite{DBLP:conf/cvpr/LiuH16} develop a two-stage deep network, where a coarse prediction map is produced and followed by a recurrent CNN to refine the details of the prediction map hierarchically and progressively. Tang and Wu~\cite{DBLP:conf/mm/TangW17} develop multiple single-scale fully convolutional networks integrated chained connections to generate saliency prediction results from coarse to fine. Recently, Hou~\etal~\cite{DBLP:journals/pami/HouCHBTT19} take full advantage of multi-level and multi-scale features extracted from fully CNNs, and introduce short connections to the skip-layer structures within the holistically-nested edge detector.

\subsection{Data Augmentation}
Data augmentation is a common method used in deep network training to deal with training data shortage. Recently, generative models including variational auto-encoder (VAE)~\cite{DBLP:conf/nips/OordKEKVG16,DBLP:conf/eccv/YanYSL16} and generative adversarial networks (GANs)~\cite{DBLP:conf/nips/GoodfellowPMXWOCB14,DBLP:conf/iccv/ZhuPIE17} are used to synthesize images similar to those in realistic scenes for data augmentation. Oord~\etal~\cite{DBLP:conf/nips/OordKEKVG16} propose a new conditional image generation method based on the Pixel-CNN structure. It can be conditioned feature vectors obtained from descriptive labels or tags, or latent embeddings created by other networks. In~\cite{DBLP:conf/eccv/YanYSL16}, a layered VAE model with disentangled latent variables is proposed to generate images from visual attributes. Besides, different from VAE, Goodfellow~\etal~\cite{DBLP:conf/nips/GoodfellowPMXWOCB14} estimate generative models via an adversarial process of two models, where the generative model captures the data distribution, and the discriminative model estimates the probability that a sample came from the training data rather than the generative model. Recently, the CycleGAN model~\cite{DBLP:conf/iccv/ZhuPIE17} is to learn the mapping between an input image and an output image in different styles.

\subsection{Domain Adaptation}
In training deep learning models, due to many factors, there exists a shift between the domains of the training and testing data that can degrade the performance. Domain adaptation uses labeled data in one source domains to apply to testing data in a target domain. Recently there have been several domain adaptation methods for visual data. In~\cite{DBLP:conf/icml/GaninL15}, the authors learn deep features such that they are not only discriminative for the main learning task on the source domain but invariant with respect to the shift between the domains. Saito~\etal~\cite{DBLP:conf/icml/SaitoUH17} propose an asymmetric tri-training method for unsupervised domain adaptation, where unlabeled samples are assigned to pseudo-labels and train neural networks as if they are true labels. In~\cite{DBLP:conf/mm/WangFCYHY18}, a novel Manifold Embedded Distribution Alignment method is proposed to learn the domain-invariant classifier with the principle of structural risk minimization while performing dynamic distribution alignment. The work of~\cite{DBLP:conf/cvpr/Chen0SDG18} adapts the Faster R-CNN~\cite{DBLP:conf/nips/RenHGS15} with both image and instance level domain adaptation components to reduce the domain discrepancy. Qi~\etal~\cite{DBLP:conf/mm/QiYX18} propose a covariant multimodal attention based multimodal domain adaptation method by adaptively fusing attended features of different modalities.

\subsection{Grocery Product Dataset}
To date, there only exists a handful of related datasets for grocery product classification~\cite{rocha2010automatic}, recognition~\cite{koubaroulis2002evaluating,DBLP:conf/cvpr/MerlerGB07,DBLP:conf/eccv/GeorgeF14,DBLP:journals/corr/JundAEB16}, segmentation~\cite{DBLP:conf/eccv/FollmannBHKU18} and tallying~\cite{DBLP:journals/corr/abs-1901-07249}.

Supermarket Produce Dataset~\cite{rocha2010automatic} includes $15$ product categories of fruit and vegetable and $2,633$ images in diverse scenes. However, this dataset is not very challenging and does not reflect the challenging aspects of real life checkout images. SOIL-47~\cite{koubaroulis2002evaluating} contains $47$ product categories, where each category has $21$ images taken from $20$ different horizontal views. Then, Grozi-120~\cite{DBLP:conf/cvpr/MerlerGB07} contains $120$ grocery product categories in natural scenes, including $676$ from the web and $11,194$ from the store. Similar to Grozi-120, Grocery Products Dataset~\cite{DBLP:conf/eccv/GeorgeF14} is proposed for grocery product recognition. It consists of $80$ grocery products comprising $8,350$ training images and $680$ testing images. The training images are downloaded from the web, and the testing images are collected in natural shelf scenario. Freiburg Groceries Dataset~\cite{DBLP:journals/corr/JundAEB16} collects $5,021$ images of $25$ grocery classes using four different smartphone cameras at various stores, apartments and offices in Freiburg, Germany, rather than collecting them from the web. Specifically, the training set consists of $4,947$ images that contains one or more instances of one class, while the testing set contains $74$ images of $37$ clutter scenes, each containing objects of multiple classes. Besides, in~\cite{DBLP:conf/eccv/FollmannBHKU18}, the MVTec D2S dataset is used for instance-aware semantic segmentation in an industrial domain. It consists of $21,000$ images of $60$ object categories with pixel-wise labels.

Different from the aforementioned datasets, the RPC dataset~\cite{DBLP:journals/corr/abs-1901-07249} is the largest scale of grocery product dataset to date, including $200$ product categories and $83,739$ images. Each image is obtained for a particular instance of a type of product with different appearances and shapes, which is divided into $17$ sub-categories, such as \texttt{puffed food}, \texttt{instant drink}, \texttt{dessert}, \texttt{gum}, \texttt{milk}, \texttt{personal hygiene} and \texttt{stationery}. Specifically, $53,739$ single-product images are taken in isolated environment as \textit{training} exemplar images. To capture multi-view of single-product images, four cameras are used to cover the top, horizontal, $30^\circ$ and $45^\circ$ views of the exemplar image on a turntable. Then, each camera takes photos every $9$ degrees when the turntable rotating. The resolution of the captured image is $2592\times1944$. Then, several random products are placed on a $80cm\times80cm$ white board, and then a camera mounted on top takes the photos with a resolution of $1800\times1800$ pixels to generate checkout images. Based on the number of products, the testing images are categorized in three difficulty levels, \ie, easy ($3\sim5$ categories and $3\sim10$ instances), medium ($5\sim8$ categories and $10\sim15$ instances), and hard ($8\sim10$ categories and $5\sim20$ instances), each containing $10,000$ images. The dataset provides three different types of annotations for the testing checkout images:
\begin{itemize}
\item \textit{shopping lists} that provide the category and count of each item in the checkout image,
\item \textit{point-level annotations} that provide the center position and the category of each item in the checkout image,
\item \textit{bounding boxes} that provide the location and category of each item.
\end{itemize}

\section{Methodology}
In this section, we present in detail our data priming scheme for data augmentation in the training of visual item tallying network for automatic check-out system. As mentioned in the Introduction, our method has two steps. The pre-augmentation step we process training images of isolated items to remove those with irrelevant poses to improve the synthesized images. In the post augmentation step, we introduce a data priming network that helps to sift synthesized images to train the visual item tallying network.

\begin{figure}[t]
\centering
\includegraphics[width=0.85\linewidth]{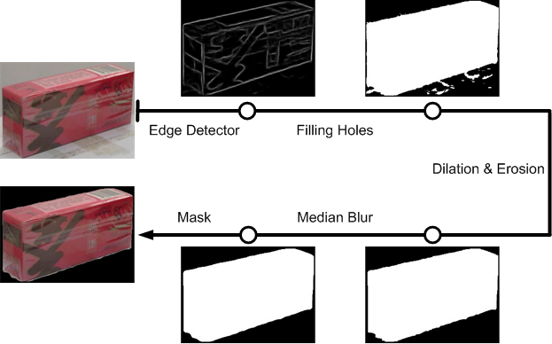}
\caption{An example of background removal by the mathematical morphology technique to generate coarse masks of items.}
\label{fig:exemplar}
\end{figure}

\subsection{Pre-augmentation Data Priming}\label{sec:exemplar}

\subsubsection{Background Removal}
Since the training images are obtained with examplar items captured on the turntable, it contains background that affects training of the visual item tallying network to focus on the object. To remove background noise, we develop a coarse-to-fine saliency based refinement method. Specifically, we first extract the contour of the object using the method of~\cite{DBLP:conf/iccv/DollarZ13}, remove the edges with the confidence score less than $0.1$, and fill the connected regions. Then other holes inside the contour are filled and small isolated regions are removed using the mathematical morphology operations such as dilation and erosion. As a last step, we use median filter to smooth the edges of the masks. A qualitative example of coarse mask generation is shown in Figure~\ref{fig:exemplar}. Given the coarse masks, we employ the saliency detection model~\cite{DBLP:journals/pami/HouCHBTT19} to extract fine masks with detailed contours of the object. The saliency model is formed by a deep neural network trained on the MSRA-B salient object database~\cite{DBLP:journals/ijcv/WangJYCHZ17}. Then, the deep neural network is fine-tuned based on the generated coarse masks of exemplars. We use these masks to extract the foreground object to use in the synthesis of testing checkout images.

\begin{figure}[t]
\centering
\includegraphics[width=0.85\linewidth]{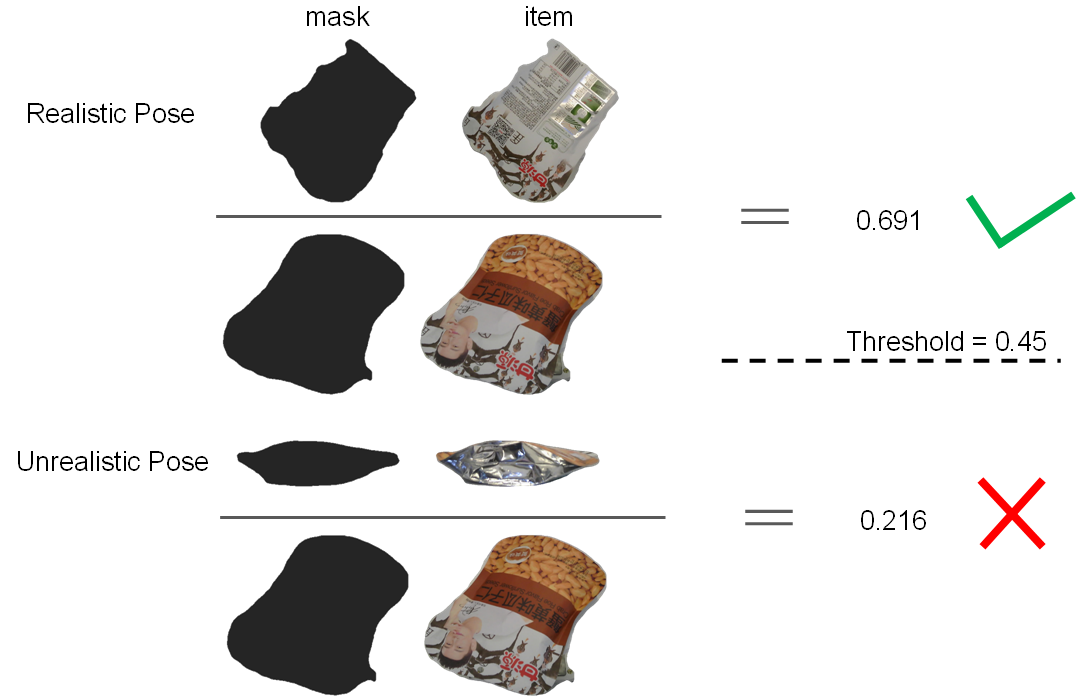}
\caption{An example of unrealistic and realistic poses from the bag-like product. We classify the item poses according to the ratio of mask area in Eq.\eqref{equ:ratio}.}
\label{fig:category}
\end{figure}

\begin{figure*}[t]
\centering
\includegraphics[width=0.9\linewidth]{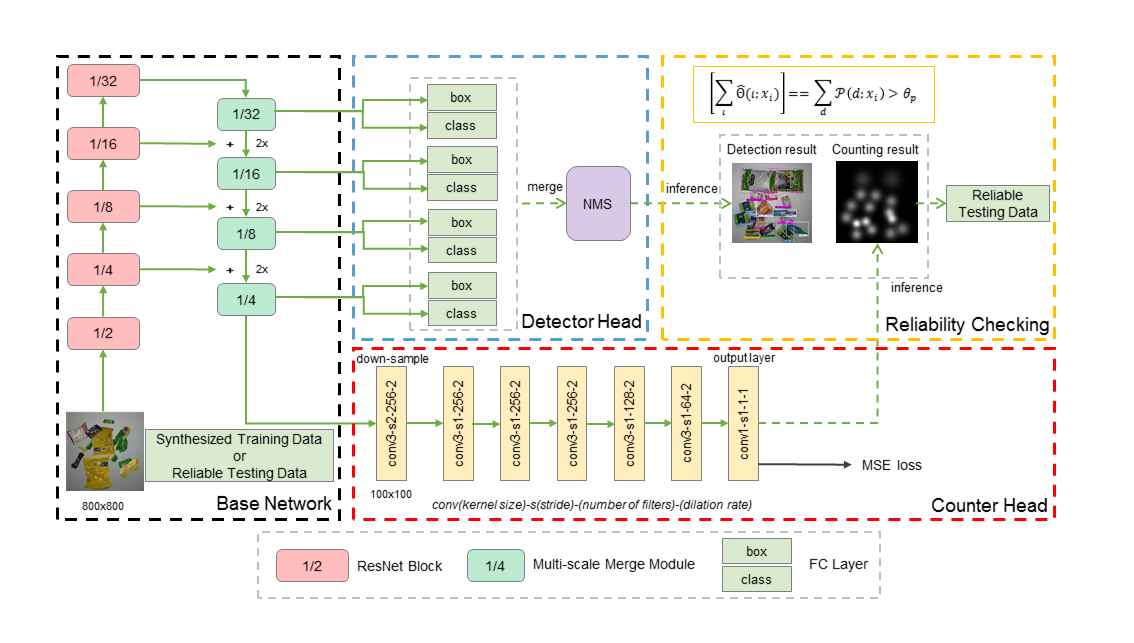}
\caption{The framework of the proposed network. The black dashed bounding box indicates the base network. The red and blue dashed bounding boxes correspond to the counter and detector heads, respectively. The orange dashed bounding box is to measure the reliability of testing data.}
\label{fig:framework}
\end{figure*}

\subsubsection{Pose Pruning}\label{sec:augmentation}
Since the testing image contains multiple objects while the training image only contains a single object, we use the segmented isolated items to create synthesized checkout images. However, not all the poses of the isolated items are viable in checkout images. For example, it is difficult to put bag-like products on the checkout table with the view from bottom to top, as shown in Figure~\ref{fig:category}. To remove these inappropriate poses of exemplars, we propose a simple metric based on the ratio of areas, \ie,
\begin{equation}
\begin{aligned}
&\mathcal{R}_{k,v} = \frac{\mathcal{A}_{k,v}}{\max_v{\mathcal{A}_{k,v}}},
\end{aligned}
\label{equ:ratio}
\end{equation}
where $\mathcal{A}_{k,v}$ is the area of the item mask captured by the $v$-th view in the $k$-th category. If the ratio is less than a pre-set threshold $\theta_m$ ($\theta_m=0.45$ in the experiment), it indicates that the area of this pose is too small to be put on the checkout table stably, \ie, unrealistic pose. Otherwise, we regard this pose as a realistic pose.

\subsubsection{Checkout Images Synthesis}
After obtaining the selected segmented items, we synthesize the checkout images using the method in~\citep{DBLP:journals/corr/abs-1901-07249}. Specifically, segmented items are randomly selected and freely placed (\ie, random angles from $0$ to $360$ and scales from $0.4$ to $0.7$) on a prepared background image such that the occlusion rate of each instance less than $50\%$. Thus the synthesized images are similar to the checkout images in terms of item placement.

\begin{figure}[t]
\centering
\includegraphics[width=0.95\linewidth]{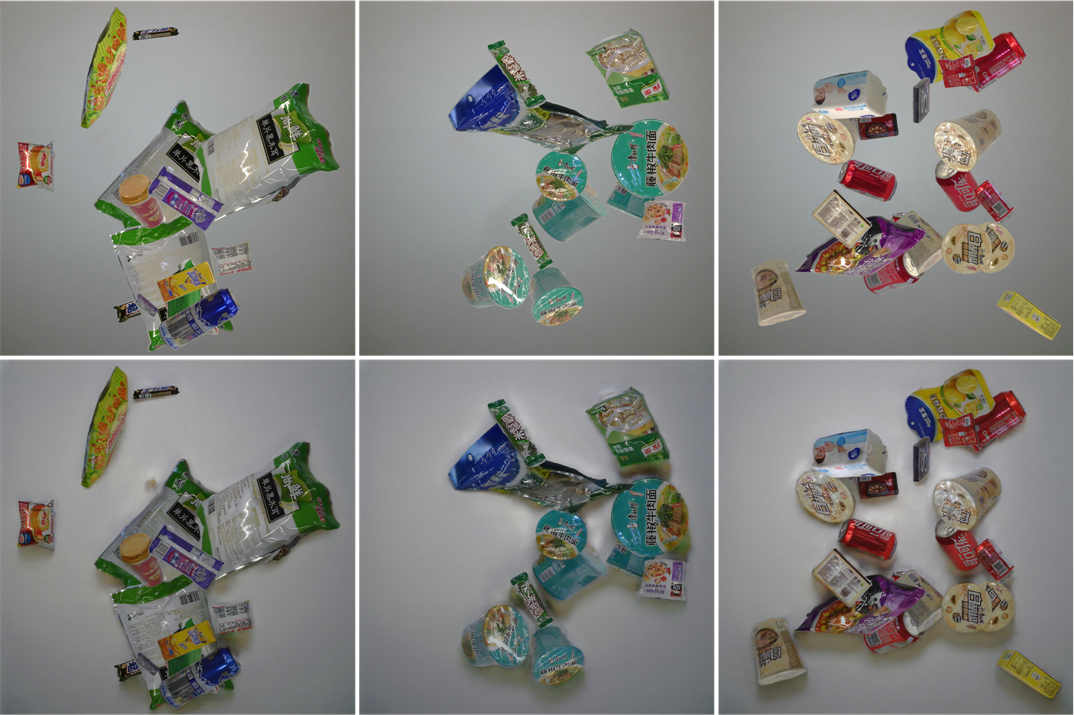}
\caption{Comparison of synthesized images (first row) and rendered images (second row) by the Cycle-GAN method~\cite{DBLP:conf/iccv/ZhuPIE17}.}
\label{fig:gan}
\end{figure}

The synthesized checkout images by random copy and paste still lack characteristics of the true testing images, so following the work in~\citep{DBLP:journals/corr/abs-1901-07249}, we use the Cycle-GAN~\cite{DBLP:conf/iccv/ZhuPIE17} to render synthesized checkout images with more realistic lighting condition and shadows, as shown in Figure~\ref{fig:gan}.

\subsection{Data Priming Network}\label{sec:priming}
We can train a deep neural network for visual item tallying using the rendered synthesized checkout images. However, the rendered images still have different characteristics with regards to the actual checkout images. To solve the problem, we propose the Data Priming Network (DPNet) to select reliable testing samples using the detection and counting collaborative learning strategy to guide the training of visual item tallying network.

\subsubsection{Network Architecture}
The goal of the visual item tallying in ACO is to predict the count and the category of items in the checkout image. To this end, we introduce a data priming network to select reliable checkout images to facilitate the training. Specifically, the data priming network consists of three components, \ie, base network $\mathcal{B}$ with counter head $\mathcal{C}$ and detector head $\mathcal{D}$, as shown in Figure~\ref{fig:framework}. $\mathcal{B}$ denotes the base network that outputs shared features among two heads, which is implemented using the ResNet-101 backbone~\cite{DBLP:conf/cvpr/HeZRS16} with Feature Pyramid Network (FPN) architecture~\cite{DBLP:conf/cvpr/LinDGHHB17}. Based on the shared features, the counter head $\mathcal{C}$ predicts the number of total instances using the predicted density map, while the detector head $\mathcal{D}$ recognizes the location and category of instances. From the last feature maps of the base network, the counter head consists of several dilated convolutional layers to extract deeper features without losing resolutions and a $1\times1$ convolutional layer as output layer, similar to~\cite{DBLP:conf/cvpr/LiZC18}. Notably, the feature maps are first down-sampled with a factor of $2$ to reduce computational complexity using a stride-$2$ dilated convolutional layer. The detector head includes fully connected layers to calculate regression and classification losses from multi-scale feature maps (\ie, $1/4,1/8,1/16,1/32$ size of the input image).

\subsubsection{Loss Function}
The loss function of the proposed network consists of terms of the counter and detector heads. For the counter head, we use the Euclidean distance to measure the difference between the ground-truth map and the estimated density map we generated. For the detector head, we use the standard cross-entropy loss for classification and smooth \text{L1} loss for regression~\cite{DBLP:conf/cvpr/LinDGHHB17}. The loss function is given as follow:
\begin{equation}
\begin{aligned}
\mathcal{L} &= \frac{1}{2\mathbb{N}}\sum_{i=1}^{\mathbb{N}}\big(\sum_{\ell}|\hat{\Theta}(\ell;x_i)-\Theta(\ell;x_i)|^2\\
&+ \lambda\sum_{d}(\mathcal{L}_\text{cls}(\hat{p}_d,p_d;x_i)+\mathbb{I}(p_d>0)\cdot\mathcal{L}_\text{reg}(\hat{t}_d,t_d;x_i))\big),
\end{aligned}
\label{equ:loss}
\end{equation}
where $x_i$ represents the input image and $\mathbb{N}$ is the batch size. $\hat{\Theta}(\ell;x_i)$ and $\Theta(\ell;x_i)$ are the estimated and ground-truth density of location $\ell$ in the input image $x_i$, respectively. Both maps are $1/8$ size of the input image. $\hat{p}_d$ and $p_d$ are the predicted and ground-truth class label of detection $d$ in the image $x_i$, including the class index of background $0$. We have $\mathbb{I}(p_d>0)=1$ if its argument is true (objects), and $0$ otherwise (background), That is, we only consider the regression loss of objects, where $\hat{t}_d$ and $t_d$ are the regression vectors representing the $4$ parameterized coordinates of the predicted and ground-truth bounding box of detection $d$ in the image $x_i$, respectively. $\lambda$ is the factor to balance the two terms.

\subsubsection{Ground-truth generation}
To train the DPNet, we need to generate ground-truth density maps. Using the center locations of extracted item masks, we generate ground-truth density maps for rendered images using the strategy in~\cite{DBLP:conf/cvpr/ZhangZCGM16}. First, we blur the center of each instance using a normalized Gaussian kernel. Then, we generate the ground-truth considering the spatial distribution of all instance in the rendered image. For the detector, both the locations and labels of instances simply come from the exemplars in the synthesized images.

\subsubsection{Detection and Counting Collaborative Learning}
We train the network using detection and counting collaborative learning, the whole procedure of which is presented in Algorithm~\ref{alg:optimization}. First, we train the entire network with the source training set. Here both the counter $\mathcal{C}$ and the detector $\mathcal{D}$ are optimized by Eq.~\eqref{equ:loss}. Then, we can select reliable testing data such that the estimated number of items by the counter head is equal to the number of detections with high confidence (we set as $\theta_p=0.95$ in the experiment) by the detection head after NMS operation, \ie,
\begin{equation}
\begin{aligned}
\big[\sum_{\ell}{\hat{\Theta}(\ell;x_i)}\big]==\sum_{d}\mathbb{I}(\mathcal{P}(d;x_i)>\theta_p),
\end{aligned}
\label{equ:condition}
\end{equation}
where $\hat{\Theta}(\ell;x_i)$ is the estimated density of location $\ell$ in the sample $x_i$ and $\big[\cdot\big]$ indicates the rounding operation. $\mathcal{P}(d;x_i)$ is the probability of detection $d$ in the sample $x_i$. $\mathbb{I}(\cdot)=1$ if its argument is true, and $0$ otherwise. Finally, after removing the counter head, the network $\mathcal{B}+\mathcal{D}$ is fine-tuned based on selected reliable testing data from target domain as the visual item tallying network.

\begin{algorithm}
\caption{Detection and Counting Collaborative Learning}\label{alg:optimization}
\begin{algorithmic}[1]
\REQUIRE rendered training data $\mathcal{S}$ with annotations, unlabelled testing data $\mathcal{T}$
\ENSURE counts and categories of items in testing data $\mathcal{T}$
\FOR{$j=1$ \texttt{to} $N_\text{iter}$}
\STATE Train the DPNet $\mathcal{B}+\mathcal{C}+\mathcal{D}$ using rendered training data $\mathcal{S}$.
\ENDFOR
\STATE Select reliable testing data $\hat{\mathcal{T}}$ based on Eq.~\eqref{equ:condition}.
\STATE Remove the counter head $\mathcal{C}$ of the DPNet to obtain the visual item tallying network.
\FOR{$j=1$ \texttt{to} $N_\text{iter}$}
\STATE Fine-tune the visual item tallying network $\mathcal{B}+\mathcal{D}$ using reliable testing data $\hat{\mathcal{T}}$.
\ENDFOR
\STATE Evaluate the visual item tallying network based on testing data $\mathcal{T}$.
\end{algorithmic}
\end{algorithm}

\section{Experiment}
We evaluate our method\footnote{Both the source codes and experimental results can be found in \url{https://isrc.iscas.ac.cn/gitlab/research/acm-mm-2019-ACO}.} on the RPC dataset~\cite{DBLP:journals/corr/abs-1901-07249} with several baseline methods.

\subsection{Implementation Details}
The propose method is implemented by PyTorch~\cite{pytorch}. The setting for the cycleGAN model is similar to that of~\cite{DBLP:conf/iccv/ZhuPIE17}. Each mini-batch consists of $2$ images on each GPU and we set the number of detections to be $256$ for each image. We use the SGD optimization algorithm to train the DPNet, and set the weight decay to be $0.0001$ and momentum is set to be $0.9$. The factor $\lambda$ in Eq.~\eqref{equ:loss} is set as $1$. For the counter head, the initial learning rate is $4\times10^{-7}$ for the first 120k iterations, which decays by a factor of $10$ for the next 40k iterations. For the detection head, the initial learning rate is $0.01$ for the first 120k iterations, which decays by a factor of $10$ for the next 40k iterations. All the experiments are conducted on a workstation with $4$ Nvidia TITAN Xp GPUs.
\begin{table*}[t]
  \centering
  \small
  \caption{Experimental results on the RPC dataset.}
  \setlength{\tabcolsep}{8.0pt}
    \begin{tabular}{|c|c||c|c|c|c|c|c|}
    \hline
    Clutter mode & Methods & cAcc ($\uparrow$) & ACD ($\downarrow$) & mCCD ($\downarrow$) & mCIoU ($\uparrow$) & mAP50 ($\uparrow$) & mmAP ($\uparrow$)\\
    \hline
    \multirow{10}[0]{*}{Easy} 
          & Single (Baseline) & 0.02\% & 7.83  & 1.09  & 4.36\% & 3.65\% & 2.04\% \\
          & Syn (Baseline) & 18.49\% & 2.58  & 0.37  & 69.33\% & 81.51\% & 56.39\% \\
          & Render (Baseline) & 63.19\% & 0.72  & 0.11  & 90.64\% & 96.21\% & 77.65\% \\
          & Syn+Render (Baseline) & 73.17\% & 0.49  & 0.07  & 93.66\% & 97.34\% & 79.01\% \\
          & Render (DPNet(w/o PP)) & 79.82\%  & 0.31 & 0.05 & 95.84\%  & 98.33\% &82.05\%  \\
          & Render (DPNet(w/o DP)) & 85.38\%  & 0.23 & 0.03 & 96.82\% & \bf{98.72\%} & 83.10\%  \\
          & Render (DPNet(w/o DPC)) & 84.46\%  & 0.23 & 0.03 & 96.92\% & 97.93\% & 83.22\% \\
          & Render (DPNet) & 89.74\% & 0.16 & 0.02 & 97.83\% & 98.52\% & 82.75\% \\
          & Syn+Render (DPNet(w/o DP)) & 86.58\% & 0.21 & 0.03 & 97.12\% & 98.62\% & \bf{83.47\%} \\
          & Syn+Render (DPNet) & \bf{90.32\%} & \bf{0.15} & \bf{0.02} & \bf{97.87\%} & 98.60\% & 83.07\% \\
    \hline
    \multirow{10}[0]{*}{Medium} 
          & Single (Baseline) & 0.00\% & 19.77  & 1.67  & 3.96\% & 2.06\% & 1.11\% \\
          & Syn (Baseline) & 6.54\% & 4.33  & 0.37  & 68.61\% & 79.72\% & 51.75\% \\
          & Render (Baseline) & 43.02\% & 1.24  & 0.11  & 90.64\% & 95.83\% & 72.53\% \\
          & Syn+Render (Baseline) & 54.69\% & 0.90  & 0.08  & 92.95\% & 96.56\% & 73.24\% \\
          & Render (DPNet(w/o PP)) & 58.76\% & 0.74 & 0.06 & 94.10\% & 97.55\% & 76.05\% \\
          & Render (DPNet(w/o DP)) & 70.90\% & 0.49 & 0.04 & 95.90\% & 98.16\% & 77.22\% \\
          & Render (DPNet(w/o DPC)) & 69.85\% & 0.50 &  0.04  & 95.95\% & 97.24\% & 77.09\% \\
          & Render (DPNet) & 77.75\% & 0.35 & 0.03 & 97.04\% & 97.92\% & 76.78\% \\
          & Syn+Render (DPNet(w/o DP)) & 73.20\%  & 0.46 &  0.04  & 96.24\% & \bf{98.19\%} & \bf{77.69\%} \\
          & Syn+Render (DPNet) & \bf{80.68\%} & \bf{0.32} & \bf{0.03} & \bf{97.38\%} & 98.07\% & 77.25\% \\
    \hline
    \multirow{10}[0]{*}{Hard} 
          & Single (Baseline) & 0.00\% & 22.61  & 1.33  & 2.06\% & 0.97\% & 0.55\% \\
          & Syn (Baseline) & 2.91\% & 5.94  & 0.34  & 70.25\% & 80.98\% & 53.11\% \\
          & Render (Baseline) & 31.01\% & 1.77  & 0.10  & 90.41\% & 95.18\% & 71.56\% \\
          & Syn+Render (Baseline) & 42.48\% & 1.28  & 0.07  & 93.06\% & 96.45\% & 72.72\% \\
          & Render (DPNet(w/o PP)) & 44.58\% & 1.20 & 0.07 & 93.25\%  & 96.86\% & 73.62\% \\
          & Render (DPNet(w/o DP)) & 56.25\% & 0.84 & 0.05 & 95.28\% & 97.67\% &74.88\%  \\
          & Render (DPNet(w/o DPC)) & 52.80\% & 0.86 & 0.05 & 95.17\% & 96.51\% & 74.77\% \\
          & Render (DPNet) & 66.35\% & 0.60 & \bf{0.03} & 96.60\% & 97.49\% & 74.67\% \\
          & Syn+Render (DPNet(w/o DP)) & 59.05\% & 0.77 & 0.04 & 95.71\% & \bf{97.77\%} & \bf{75.45\%} \\
          & Syn+Render (DPNet) & \bf{70.76\%} & \bf{0.53} & \bf{0.03} & \bf{97.04\%} & 97.76\% & 74.95\% \\
    \hline
    \multirow{10}[0]{*}{Averaged} 
          & Single (Baseline) & 0.01\% & 12.84  & 1.06  & 2.14\% & 1.83\% & 1.01\% \\
          & Syn (Baseline)  & 9.27\% & 4.27  & 0.35  & 69.65\% & 80.66\% & 53.08\% \\
          & Render (Baseline) & 45.60\% & 1.25  & 0.10  & 90.58\% & 95.50\% & 72.76\% \\
          & Syn+Render (Baseline) & 56.68\% & 0.89  & 0.07  & 93.19\% & 96.57\% & 73.83\% \\
          & Render (DPNet(w/o PP)) & 60.98\%  & 0.75 & 0.06 & 94.05\%  & 97.29\%  & 75.89\%\\
          & Render (DPNet(w/o DP)) & 70.80\% & 0.52 & 0.04 & 95.86\% & 97.93\% & 77.07\% \\
          & Render (DPNet(w/o DPC)) & 69.03\% & 0.53 & 0.04 & 95.82\% & 96.96\% & 77.09\% \\
          & Render (DPNet) & 77.91\% & 0.37 & \bf{0.03} & 97.01\% & 97.74\% & 76.80\% \\
          & Syn+Render (DPNet(w/o DP)) & 72.83\% & 0.48 & 0.04 & 96.17\% & \bf{97.94\%} & \bf{77.56\%} \\
          & Syn+Render (DPNet) & \bf{80.51\%} & \bf{0.34} & \bf{0.03} & \bf{97.33\%} & 97.91\% & 77.04\% \\
    \hline
    \end{tabular}%
  \label{tab:experiment}%
\end{table*}%

\begin{figure*}[t]
\centering
\includegraphics[width=0.85\linewidth]{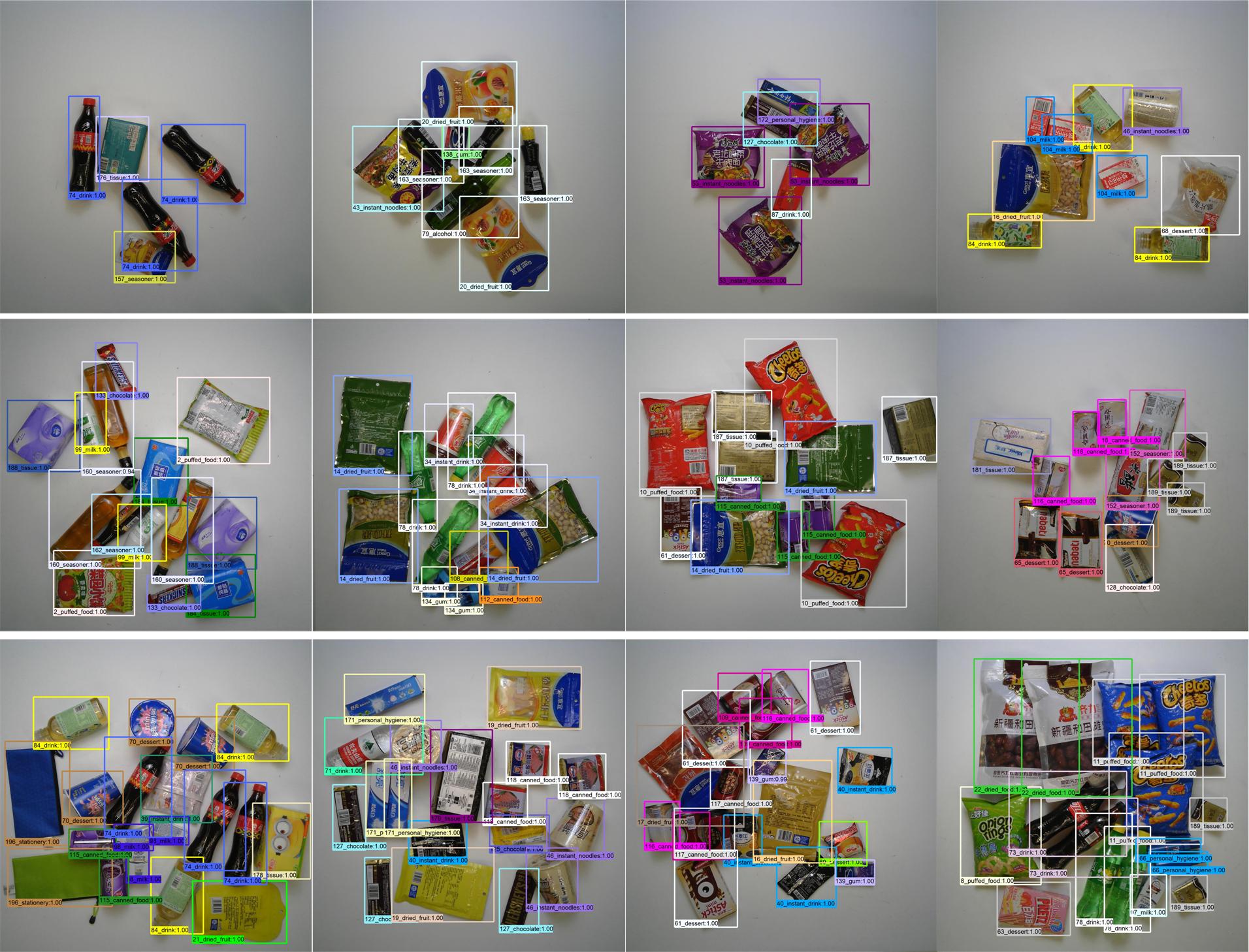}
\caption{Results of our method for easy, medium, and hard modes (from top to down). Different color bounding boxes correspond to the predictions with the item categories and the confidence scores. Best view in color.}
\label{fig:example}
\end{figure*}

\subsection{Evaluation protocol}
To evaluate the performance of the proposed method, we use several metrics following~\cite{DBLP:journals/corr/abs-1901-07249}. First, the counting error for a specific category in an image is defined as
\begin{equation}
\begin{aligned}
\text{CD}_{i,k} = |P_{i,k}-\text{GT}_{i,k}|,
\end{aligned}
\label{equ:cdi}
\end{equation}
where $P_{i,k}$ and $\text{GT}_{i,k}$ indicates the predicted count and ground-truth item number of the $k$-th category in the $i$-th image, respectively. To measure the error over all $K$ categories for the $i$-th image is calculated as
\begin{equation}
\begin{aligned}
\text{CD}_{i} = \sum_{k=1}^K{\text{CD}_{i,k}},
\end{aligned}
\label{equ:cd}
\end{equation}

\subsubsection{Checkout Accuracy}
Checkout Accuracy (\text{cAcc}) is the primary metric for ranking in the ACO task~\cite{DBLP:journals/corr/abs-1901-07249}, which is the accuracy when the complete product list is predicted correctly. It is calculated as
\begin{equation}
\begin{aligned}
&\text{cAcc} = \frac{\sum_{i=1}^N\delta(\sum_{k=1}^K{\text{CD}_{i,k}}=0)}{N},
\end{aligned}
\label{equ:cacc}
\end{equation}
where $\delta(\cdot)=1$ if its argument is true, and $0$ otherwise. The range of the \text{cAcc} score is from $0$ to $1$. For example, if $\sum_{k=1}^K{\text{CD}_{i,k}}=0$, all items are accurately predicted, \ie, $\text{cAcc}=1$.

\subsubsection{Mean Category Intersection of Union}
Mean Category Intersection of Union (\text{mCIoU}) measures the compatibility between the predicted product list and ground-truth. It is defined as
\begin{equation}
\begin{aligned}
&\text{mCIoU} = \frac{1}{K}\sum_{k=1}^K\frac{\sum_{i=1}^N{\min(\text{GT}_{i,k},P_{i,k})}}{\sum_{i=1}^N{\max(\text{GT}_{i,k},P_{i,k})}}.
\end{aligned}
\label{equ:mciou}
\end{equation}
The range of the \text{mCIoU} score is from $0$ to $1$.

\subsubsection{Average Counting Distance}
Different from \text{cAcc} focusing on the counting error, Average Counting Distance (\text{ACD}) indicates the average number of counting errors for each image, \ie,
\begin{equation}
\begin{aligned}
&\text{ACD} = \frac{1}{N}\sum_{i=1}^N\sum_{k=1}^K{\text{CD}_{i,k}}.
\end{aligned}
\label{equ:acd}
\end{equation}

\subsubsection{Mean Category Counting Distance}
Moreover, the Mean Category Counting Distance (\text{mCCD}) is used to calculate the average ratio of counting errors for each category, \ie,
\begin{equation}
\begin{aligned}
&\text{mCCD} = \frac{1}{K}\sum_{k=1}^K\frac{\sum_{i=1}^N{\text{CD}_{i,k}}}{\sum_{i=1}^N{\text{GT}_{i,k}}}.
\end{aligned}
\label{equ:mccd}
\end{equation}

\subsubsection{Mean Average Precision}
On the other hand, according to the evaluation protocols in MS COCO \cite{DBLP:conf/eccv/LinMBHPRDZ14} and the ILSVRC 2015 challenge \cite{DBLP:journals/ijcv/RussakovskyDSKS15}, we use the mean Average Precision (mAP) metrics (\ie, \text{mAP50} and \text{mmAP}) to evaluate the performance of the detector. Specifically, \text{mAP50} is computed at the single Intersection over Union (IoU) threshold $0.50$ over all item categories, while \text{mmAP} is computed by averaging over all $10$ IoU thresholds (\ie, in the interval $[0.50,0.95]$ in steps of $0.05$) of all item categories.

\subsection{Baseline Solutions}
The authors of~\cite{DBLP:journals/corr/abs-1901-07249} provide four baselines for comparison. Specifically, a detector is trained to recognize the items based on the following four kinds of training data.
\begin{itemize}
\item \textit{Single}. We train the FPN detector~\cite{DBLP:conf/cvpr/LinDGHHB17} using training images of isolated items based on the bounding box annotations.
\item \textit{Syn}. We copy and paste the segmented isolated items to create $100,000$ synthesized checkout images for detector training. To segment these items, we employ a salience based object segmentation approach~\cite{DBLP:journals/tip/HuWZL16} with Conditional Random Fields (CRF)~\cite{DBLP:conf/nips/KrahenbuhlK11} for mask refinement of item to remove the background noise.
\item \textit{Render}. To reduce domain gap, we employ Cycle-GAN~\cite{DBLP:conf/iccv/ZhuPIE17} to translate the synthesized images into the checkout image domain for detector training, resulting in more realistic render images.
\item \textit{Syn+Render}. We train the detector based on both synthesized and rendered images.
\end{itemize}

\subsection{Experimental Results and Analysis}
The performance compared with $4$ baseline methods are presented in Table~\ref{tab:experiment}. More visual examples for different difficulty levels are shown in Figure~\ref{fig:example}. The \textit{Single} method fails in ACO task because of the huge gap between the exemplars and the checkout images. By combining segmented items into synthesized checkout images, the checkout accuracy is improved from $0.01\%$ to $9.27\%$ in averaged level. Moreover, significant boost is achieved by training the detector on rendered images. This is because the GAN method can mimic the realistic checkout images in lighting conditions or shadow patterns effectively. Compared to the aforementioned \textit{Render} baseline method (\ie, $45.60\%$ \text{cAcc} score), our DPNet method achieves $77.91\%$ \text{cAcc} score in averaged level only training on rendered images. Given the \textit{Syn+Render} data, the checkout accuracy is further improved by $17.15\%$, $25.99\%$, $28.28\%$ for easy, medium and hard level respectively compared with the \textit{Syn+Render} baseline method. This indicates the effectiveness of our approach.

\subsection{Ablation Study}
We further perform experiments to study the effect of different modules of the proposed method by construct three variants, \ie, DPNet(w/o DPC), DPNet(w/o DP) and DPNet(w/o PP). DPNet(w/o DPC) indicates that the DPNet removes the counter head to select reliable testing data. In this way, the reliability checking condition in Eq.~\eqref{equ:condition} is rewritten as $\sum_{d}\mathbb{I}(\mathcal{P}(d;x_i)>\theta_p)\geq3$, because the least number of items in the checkout image is $3$ (easy mode). DPNet(w/o DP) indicates that we do not use the DPNet for domain adaptation, \ie, the detector is trained based on the rendered data. DPNet(w/o PP) denotes the method that further removes the pose pruning module from DPNet(w/o DP). For fair comparison, we use the same parameter settings and input size in evaluation. We choose all testing checkout images to conduct the experiments.

\subsubsection{Effectiveness of Background Removal}
The \textit{Render} baseline method uses the Saliency~\cite{DBLP:journals/tip/HuWZL16}+CRF~\cite{DBLP:conf/nips/KrahenbuhlK11} model to obtain the masks of exemplars. As presented in Table~\ref{tab:experiment}, our DPNet(w/o PP) method achieves better performance, \ie, $60.98\%$ vs. $45.60\%$ checkout accuracy based on the rendered data. This may be attributed to better segmentation results by our DPNet(w/o PP) method using coarse-to-fine strategy.

\subsubsection{Effectiveness of Pose Pruning}
If we remove the pose pruning module, the DPNet(w/o PP) method decreases $9.82\%$ in terms of checkout accuracy ($60.98\%$ vs. $70.80\%$). This noticeable performance drop validates the importance of the pose pruning module to remove the synthesized images including the items with unrealistic poses (see Figure~\ref{fig:category}).

\subsubsection{Effectiveness of Detection and Counting Collaborative Learning}
From Table~\ref{tab:experiment}, our proposed DPNet achieves better results than its variant DPNet(w/o DP). The increase in checkout accuracy indicates that the data priming method adapts the data from source domain to that from target domain effectively. Besides, DPNet(w/o DPC) performs even slightly inferior than DPNet(w/o DP), \ie, ($69.03\%$ vs. $70.80\%$). It is not confident to determine reliable testing data only based on the detection head, resulting in much unreliable testing data ($34.9\%$ of selected testing data). On the contrary, we can select $90.7\%$ correct reliable testing data based on the proposed DPNet with both counter and detection heads. 

\begin{figure}[t]
\centering
\includegraphics[width=0.95\linewidth]{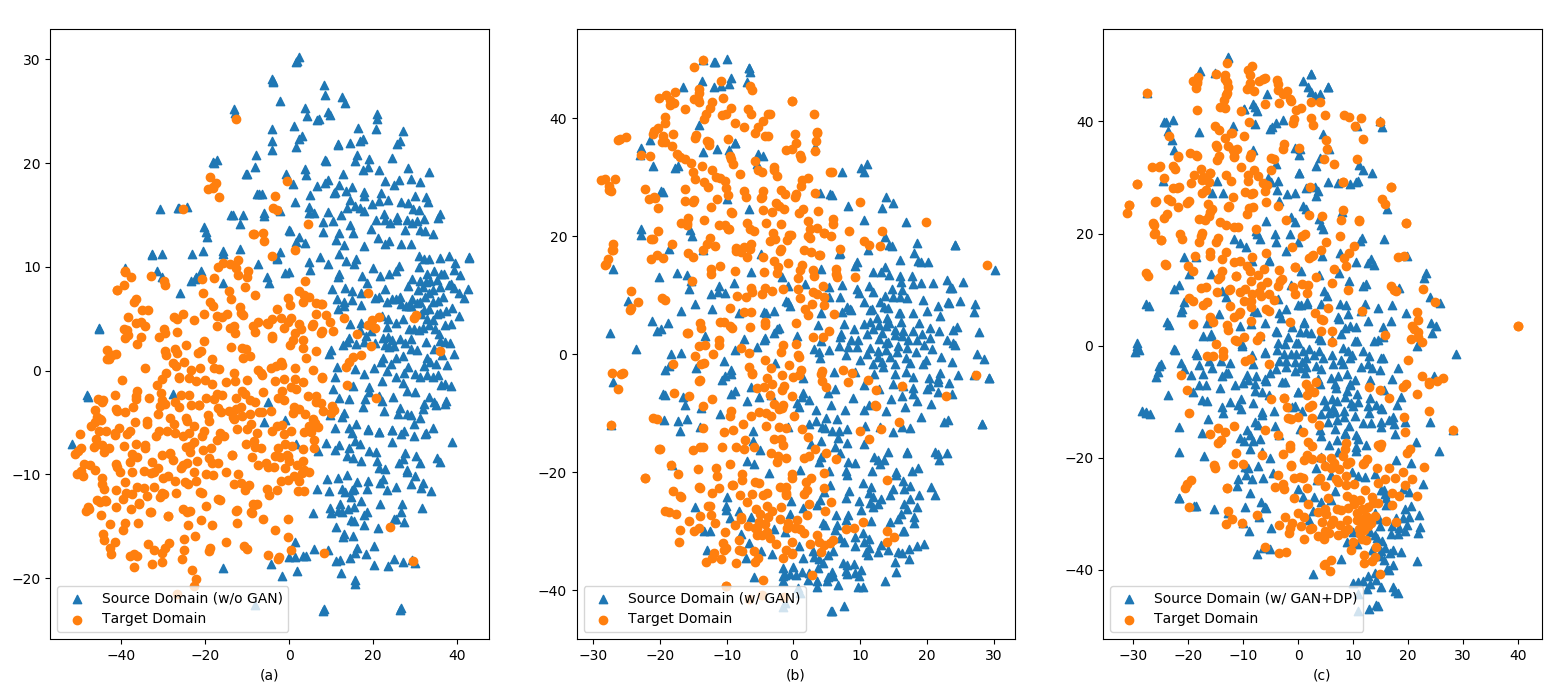}
\caption{Comparison of the distribution of source and target domain based on the detector trained on (a) synthesized data; (b) rendered data; (c) reliable testing data.}
\label{fig:adaptation}
\end{figure}

To visualize the distribution of source and target domains, we first train our network with the ResNet-101 backbone and then calculate the features of $500$ images randomly selected from the two domains using the last block of backbone. For Figure~\ref{fig:adaptation}(a), the network is trained on synthesized images. For Figure~\ref{fig:adaptation}(b), the network is trained on rendered images. For Figure~\ref{fig:adaptation}(c), the network is trained on rendered image and then finetuned on reliable testing images. Finally, we embed high-dimensional features of each domain for visualization in a low-dimensional space of two dimensions using the t-Distributed Stochastic Neighbor Embedding (t-SNE) technique~\cite{DBLP:journals/jmlr/Maaten14}.

\subsubsection{Effectiveness of \textit{Syn+Render}}
Similar to the trend in the baseline methods ($45.60\%$ \text{cAcc} of \textit{Render} (baseline) vs. $56.68\%$ \text{cAcc} of \textit{Syn+Render} (baseline)), the performance is constantly improved when training on both synthesized and rendered data. Specifically, Syn+Render (DPNet) achieves $80.51\%$ \text{cAcc} score compared to $77.91\%$ \text{cAcc} score of the Render (DPNet) configuration.

\section{Conclusion}
In this paper, we propose a new data priming network to deal with automatic checkout. Different from the previous domain adaptation methods, we construct both counter and detector heads to measure the reliability of testing images for the target domain. Then, the detector of the target branch can learn target-discriminative representation based on the reliable testing samples using detection and counting collaborative learning, resulting in robust performance. The experiment on the RPC dataset shows that our method surpasses the previous baseline methods significantly by more than $20\%$ checkout accuracy in the averaged level. For future works, we would like to further study other potential options for the data priming network, including heads of other types of attributes.

\section{Acknowledgments}
This work is partially supported the National Natural Science Foundation of China under Grant 61807033, and partially supported by US Natural Science Foundation under Grant IIS1816227.

\bibliographystyle{ACM-Reference-Format}
\bibliography{acmart}

\end{document}